\documentclass[10pt,letterpaper]{article}
\usepackage{cvpr}
\usepackage{times}
\usepackage{epsfig}
\usepackage{graphicx}
\usepackage{amsmath}
\usepackage{amssymb}
\usepackage[rightcaption]{sidecap}
\usepackage{algorithmic}
\usepackage{amsopn}
\usepackage{tabularx, booktabs}
\usepackage{mathtools}
\usepackage[ruled,vlined,linesnumbered,lined,boxed,commentsnumbered]{algorithm2e}
\newtheorem{mydef}{Definition}
\usepackage{mathtools}
\usepackage{tabularx, booktabs}
\usepackage{subfig}
\usepackage{latexsym}
\usepackage{comment}
\newtheorem{Lemma}{Lemma}

\cvprfinalcopy 

\begin{document}

\title{Spectral Embedding of Graph Networks}

\author{Shay Deutsch\\
Department of Mathematics \\
$\& $ Computer Science \\
University of California Los Angeles\\
{\tt\small shaydeu@math.ucla.edu}
\and 
Stefano Soatto\\
Department of Computer Science \\
University of California Los Angeles\\
{\tt\small soatto@cs.ucla.edu }
}

\maketitle

\begin{abstract}
We introduce an unsupervised graph embedding that trades off local node similarity and connectivity, and global structure. The embedding is based on a generalized graph Laplacian, whose eigenvectors compactly capture both network structure and neighborhood proximity in a single representation. The  key idea is to transform the given graph into one whose weights measure the centrality of an edge by the fraction of the number of shortest paths that pass through that edge, and employ its spectral proprieties in the representation. 
Testing the resulting graph network representation shows significant improvement over the sate of the art in data analysis tasks including social networks and material science. We also test our method on node classification from the human-SARS CoV-2 protein-protein interactome.  
\end{abstract}

\section{Introduction}

\begin{figure}
\centering

\includegraphics[width=0.8\columnwidth]{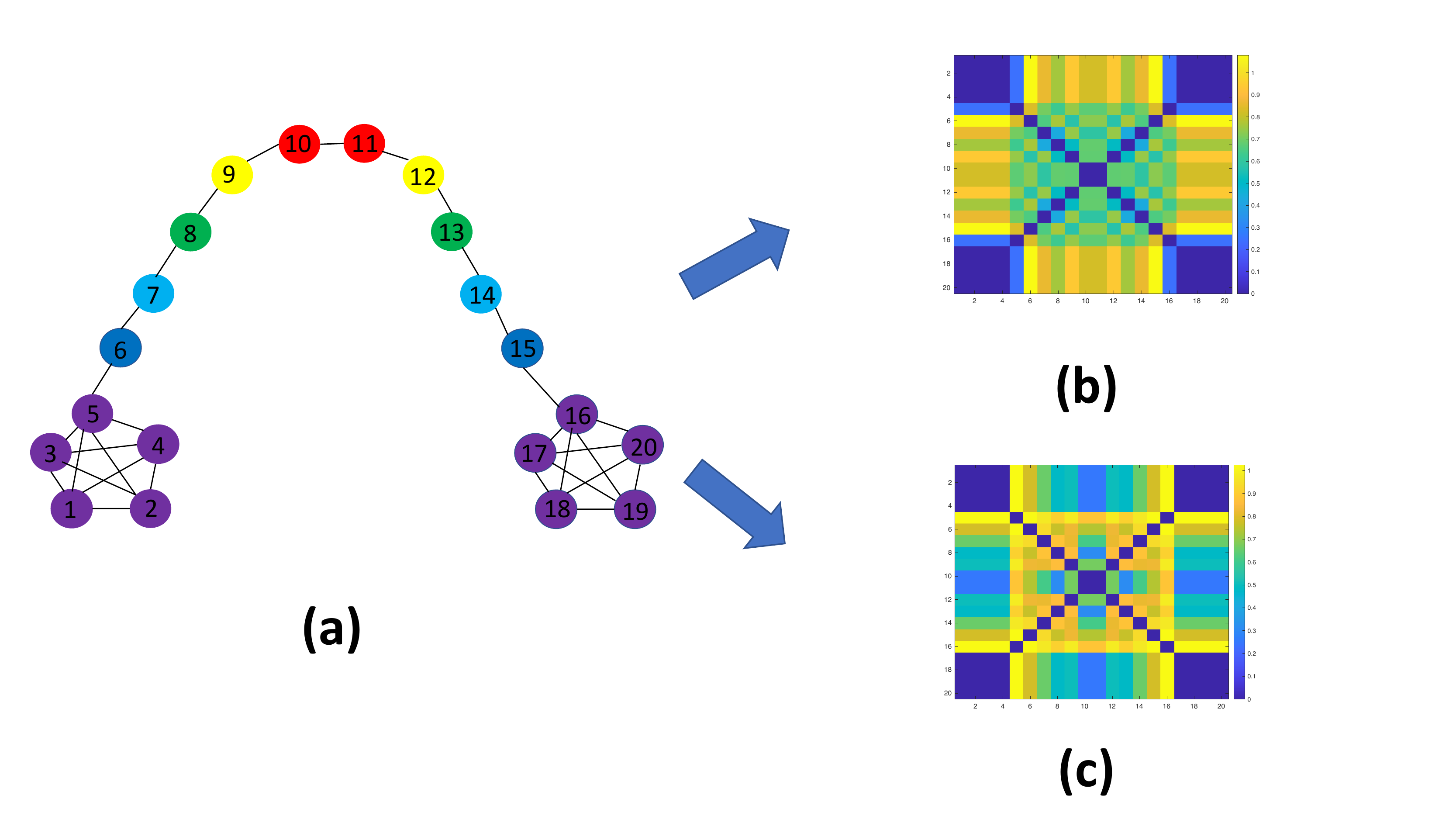}

\caption{In the Barbell graph network (Fig.(a)) we can observe two ways to organize nodes in the graph; one which is based on local similarity, and another that is based on similar network neighborhood, which is associated with node's structural role. Heat maps illustrating distances between the nodes in the graph embeddings space using the unweighted Laplacian in (b) and using the proposed generalized Laplacian in (c). The distances between the embedded features using our approach captures both structural and proximity similarity between nodes.}
\label{fig:barbell_graph_illustrate}
\end{figure}

Consider the ``barbell'' graph in Fig. \ref{fig:barbell_graph_illustrate}(a). There are two qualitatively different kinds of nodes: Those on the handle (6 through 15), and those at its ends (1-5 and 16-20). Some nodes play a  structural role, such as 5 and 16 that join the ends to the handle. Our goal is to design a representation of the graph that can capture both the local similarity between adjacent nodes, as well as the non-local similarity of distant nodes based on their structural properties. Such a representation would empower network topology analysis in fields as diverse as material science, social science, biology and commerce. The key contribution of this paper is an unsupervised approach to learn an embedding function associated with each node, that trades off local and structural similarity. At the core is a method that employs a set of two basis functions to represent the graph structure.

We propose to employ basis functions that allow to compactly capture local and structural similarity using the {\em edge betweeness centrality graph}. For a graph ${\cal G}= (\cal V, W)$ with vertices in the set $\cal V$ and edge weights in $\cal W$, we measure the {\em centrality} of an edge by the fraction of the number of shortest paths that pass through that edge, called {\em edge betweenness centrality} (EBC). From the betweenness centrality of the edges of ${\cal G}$, we construct a modified graph with the same connectivity of ${\cal G}$, but edges weighted by their EBC value. We call the resulting graph the edge betweeness centrality graph (BCG), and indicate it with ${\cal G}^{\rm BE}$. 
Note that the EBC computes the importance of an edge in the network in terms of how often it is on shortest paths between origin and destination nodes. We use the spectral proprieties of the BCG to define our embedding in a way that captures non-local variations. 

We consider two properties of the BCG, which are not necessarily manifest in the original graph ${\cal G}$: First, edges with high EBC tend to distribute almost uniformly in different parts of real-world networks. Second, the EBC can take a broad range of values across such networks. 
We use these properties to construct graph embeddings, and show that they are informative of the structure of the graph.

The embedding we propose captures both local as well as structural properties that each node can exhibit simultaneously.  
While structural information is localized, in the sense of being associated to a particular node, it is not local, as it depends on properties of nodes beyond its neighbors. 

Our approach is reminiscent of uncertainty principles developed in graph harmonic analysis~\cite{Agaskar_2013}, that explore to what extent signals can be represented simultaneously in the vertex and spectral domains. We extend this approach to characterize the relation between the spectral and non-local graph domain spread of signals defined on the nodes. We do so using a generalized Laplacian whose node embeddings simultaneously capture local and structural properties.

One can employ the generalized Laplacian derived from our approach, by applying  off-the-shelf techniques to generate graph representations that capture complex patterns in the graph networks.
While in general there is no single ``true" universal way to capture embedding distances between nodes on the graph, our approach allows flexibility in generating embeddings which can be useful for several tasks. We show our method's viability in a number of unsupervised learning tasks including node classification in social networks, node in protein-to-protein interaction (PPI) networks which include the human-SARS- CoV-2 protein-protein interactome, clustering and forecasting failure of edges in material science. 

\section{Related Work}\label{sec:RelatedWork}

We employ BCG as a tool to rank the importance of edges in the graph network given an adjacency matrix. 

Graph centrality based measurements, such as vertex centrality, eigenvector centrality \cite{Freeman1977Set}, and edge betweeness centrality \cite{Girvan2002Community} have been widely used in network analysis and its diverse applications \cite{Berthier_2019,CUCURINGU_2016}. 
Feature learning in a graph  $\cal{G}=(V,W)$, which aims to learn a function $\Phi:\cal V \rightarrow$  $ \mathbb{R}^{m} $ from the graph nodes into a vector space $ \mathbb{R}^{m}$, has recently gained considerable interest in network analysis \cite{grover2016node2vec,Ribeiro_2017, Deepwalk}. Graph convolutional networks  \cite{Kipf:2016, NIPS2016CNNG, velickovic2018graph} are among the most successful for graph based feature learning, combining the expressiveness of neural networks with graph structures. 

Our work is related to  uncertainty principles in graphs \cite{Agaskar_2013} and studied in later works \cite{Teke_U_Principles}. \cite{Agaskar_2013} suggests extending traditional uncertainty principles in signal processing to more general domains such as irregular and unweighted graphs \cite{GSP_irregular}. \cite{Agaskar_2013} provides definitions of graph and spectral spreads, where the graph spread is defined for an \textit{arbitrary} vertex and studies to what extend a signal can be localized both in the graph and spectral domains.  

Other related work includes different approaches in manifold learning \cite{LLE}, \cite{Belkin:2003}, \cite{Tenenbaum00}, manifold regularization \cite{MFD}, \cite{ZSLwavelets}, \cite{ijcai15}, \cite{SSVM15}
\cite{DeutschMS19}, \cite{DeutschM17}, \cite{ZSL_Isoperimteric},
graph diffusion kernels \cite{CoifmanDM} and kernel methods widely used in computer graphics \cite{AubrySC11, HKS} for shape detection. Most methods assume that signals defined over the graph nodes exhibit some level of smoothness with respect to the connectivity of the nodes in the graph and therefore are biased to capture local similarity.

\section{Preliminaries and Definitions}
\label{sec:Preliminaries}

Consider an undirected, weighted graph $\cal{G}=(V,W)$ with nodes $\cal V $ $ = \left \{ 1, 2, . . . , N \right \} $  and edges $\mathbf{\cal{W}} = \left \{ (w_{ij}) (i, j) \in \cal{V}  \right \} $ where $w_{ij}$ denotes the weight of an edge between nodes $i$ and $j$. The degree $d(i)$ of a node is the sum of the weights of edges connected to it. 
 The combinatorial graph Laplacian is denoted by $\mathbf{L}$, and defined as $\mathbf{L}=\mathbf{D}-\mathbf{W}$, with  $\mathbf{D}$ the diagonal degree matrix with entries $d_{ii}=d(i)$. The eigenvalues and eigenvectors of $\mathbf{L}$ are denoted by $\lambda_1,\ldots,\lambda_N $ and $\mathbf{\phi}_{1},\ldots,\phi_N$, respectively.

\subsubsection{The Edge Betweeness Centrality Graph (BCG)}

The edge betweeness centrality (EBC) is defined as 
\begin{equation}
w_{i,j}^{BE} =
\sum_{s \neq t} \frac{ \sigma_{st} (w_{ij})} {\sigma_{st}}
\end{equation} where $\sigma_{st}$ is the number of the shortest distance paths from node $s$ to node $t$ and $\sigma_{st} (w_{ij}) $ is the number of those paths that includes the edge $w_{ij}$.
Next we defined the BCG $\cal G^{BW} = (\cal V, W^{BE}) $ which modifies the edges $w_{ij}$ in $\cal G$ to the EBC weights $w_{i,j}^{BE}$ and shares the same connectivity as $\cal G$. Formally the similarity matrix representing connectivity between nodes in the graph $\cal G^{BW} = (\cal V, W^{BE}) $ is given by 
\begin{equation}
\label{Aff_Bet_graph}
{ ( {\mathbf{W})_{i,j}}^{BE}} = \left\{\begin{matrix}
\sum_{s \neq t} \frac{ \sigma_{st} (w_{ij})} {\sigma_{st}} & \mbox{if}\,\, i\sim~j \, \, \mbox{in} \,\,\,\cal G \\ 
0 & \mbox{else.}
\end{matrix}\right.
\end{equation}
Note that while the graph $\cal G^{BW} $ shares the same connectivity as $\cal G$, the spectral representation given by the eigensystem of the graph Laplacian $\mathbf{L}^{BE}$, the eigenvalues and their associated eigenvectors  $\lambda^{BE}_{1},\ldots,\lambda^{BE}_{N} $ and $\mathbf{\phi}^{BE}_{1},\ldots,\phi^{BE}_{N}$, are rather different.  
Importantly, the eigenvectors $\mathbf{\phi}^{BE}_{1},\ldots,\phi^{BE}_{N}$ provide a different realization of the graph structure in comparison to the eigenvectors of an unweighted graph, that is captured by a different diffusion process around each vertex.

\subsubsection{Total Vertex and Spectral Spreads }
\label{sec:vertex_spectral_spread}

The vertex spread of a general weighted and undirected graph is defined as follows:  

\begin{mydef}{(Total vertex domain spread)}\\
The global spread of a non zero signal $\mathbf{x}  \in l^{2}(\cal G)$ with respect to a matrix $\mathbf{W}$ (corresponding to an arbitrary affinity matrix) is defined as 
\begin{equation}
 \mathbf{g}_{\mathbf{W}}(\mathbf{x}) =  \frac{1}{ || \mathbf{x} ||^{2}}   \sum_{ i \sim j}   w_{ij}  x(i) x(j) =   \frac{1}{ || \mathbf{x} ||^{2}} \mathbf{x}^{T}\mathbf{W}\mathbf{x}
\end{equation}where $i \sim j$ corresponds to vertices connected in the similarity graph $\mathbf{W}$.  

\end{mydef}

\begin{mydef}{(Spectral spread)  \cite{Agaskar_2013}}\\
The spectral spread of a non zero signal $\mathbf{x}  \in l^{2}(\cal G)$ with respect to a similarity affinity matrix $\mathbf{W}$ is defined as 
\begin{equation}
\label{spectral_spread}
\mathbf{g}_{\mathbf{L}}(\mathbf{x}) =  \frac{1}{ || \mathbf{x} ||^{2}}   \sum_{ l=1 }^{N}  \lambda_{l} | {\hat{x}}(l)|^{2}=    
 \frac{1}{ || \mathbf{x} ||^{2}} \mathbf{x}^{T}\mathbf{L}\mathbf{x}
\end{equation}
where 
\begin{equation}
 {\hat{x}}(l) =  \sum_{i} x(i) \phi^{*}_{l}(i)
\end{equation}is the graph Fourier transform of the signal  $\mathbf{x}  \in l^{2}(\cal G)$ with respect to the eigenvalue $\lambda_{l}$.  
\end{mydef}

To distinguish between the vertex and spectral spreads of the input similarity matrix $\mathbf{W}$ and the edge betweeness centrality graph  $\mathbf{W}^{BE}$, we denote the total vertex domain and spectral spreads of the former as $ \mathbf{g}_{\mathbf{W} }(\mathbf{x}),  \mathbf{g}_{\mathbf{L}} (\mathbf{x}) $ while the latter is denoted as $ \mathbf{g}_{\mathbf{W}^{BE} }(\mathbf{x}),  \mathbf{g}_{\mathbf{L}^{BE}} (\mathbf{x}) $, respectively. 

\section{Proposed Node Embeddings}
\label{Properties_Spread}

To find node embeddings that represent the local and structural properties in the graph, we utilize the revisited definitions of  vertex and spectral spreads of the graph provided in the previous section. We focus on the spreads of the BCG $\cal G^{BW} = (V, \mathbf{W}^{BE}) $, and embeddings using $\cal G^{BE}$ that delineates a trade-off between local connectivity and structural role properties of the graph's nodes.
The trade-off will be realized by characterizing the region enclosing all possible pairs of vertex and spectral spreads of the BCG $\cal G^{BW} = (V, \mathbf{W}^{BE})$. Specifically, searching for the lower boundary of the feasibility region of the vertex and spectral spreads of  $\cal G^{BW} = (V, \mathbf{W}^{BE}) $ is shown to yield a generalized eigenvalue problem, whose corresponding eigenvectors producing a representation which trade-off between local and structural node similarity. 

The induced edge rankings from $\cal G = (V, \mathbf{W})$ in $\cal G^{BW} = (V, \mathbf{W}^{BE})$ results in changes of the associated eigenvalues and eigenvectors, which now reflects the communication network using the EBC measurement.

 \subsection{The Graph-Spectral Feasibility Domain}   
 We derive the proposed graph representation from the curve enclosing the feasibility domain of vertex and spectral spreads of the BCG. The definition below is given for a general weighted graphs. 

\begin{mydef}{Feasibility domain of $\cal G = (V, \mathbf{W})$}\\
The feasibility domain of the spectral and graph spreads corresponding to the BCG is 
\begin{equation}
\label{feasiblity_doamin}
\mathbb{D}_{ ( s_{\mathbf{L} },  s_{\mathbf{W} }   )  } = \left \{  (s_{\mathbf{L} }, s_{\mathbf{W} })  | \, g_{\mathbf{L} }(\mathbf{x})  = s_{\mathbf{L} }  ,  g_{  \mathbf{W} }(\mathbf{x})  =   s_{ \mathbf{W}   } , \\
\, \mathbf{x} \in l^{2}(\cal G)  \right \}
\end{equation} 
\end{mydef}
Note that $\mathbb{D}_{ ( s_{\mathbf{L} },  s_{\mathbf{W}}  )  }$ contains all possible values $( s_{\mathbf{L} }, s_{\mathbf{W}})$ that can be obtained on $\cal G$ with respect to a non zero signal $\mathbf{x} \in l^{2}(\cal G) $.  \\
 \subsection{Graph Differentiation  }
The lower boundary of the curve enclosing the feasibility domain of the spreads $\left (\mathbf{s}_{\mathbf{L}^{BE} }(\mathbf{x}) , \mathbf{s}_{\mathbf{W}^{BE} }(\mathbf{x})  \right )$, with respect to a unit norm vector $\mathbf{x} \in l^{2}(\cal G) $ is defined as 
\begin{equation}
\label{Feasiblility_Domain}
\left\{\begin{matrix}
\Gamma_{ \mathbf{s}_{\mathbf{W}^{BE}  } } (\mathbf{s}_{\mathbf{L}^{BE} }) =  \underset{\mathbf{x}}{\mbox{min}} \, \,
 g_{  \mathbf{W}^{BE} }   (\mathbf{x})  & \\ 
\mbox{s.t} \,  \,   g_{  \mathbf{L}^{BE} }(\mathbf{x})  =  \mathbf{s}_{\mathbf{L}^{BE} }  \, \mbox{and}  \, \, \mathbf{x}^{T}\mathbf{x} =1 & 
\end{matrix}\right.
\end{equation}
To solve (\ref{Feasiblility_Domain}), we use the Lagrangian 
\begin{equation}
L(\mathbf{x}, \beta , \lambda)=  \, \,\mathbf{x}^{T}  \mathbf{W}^{BE} \mathbf{x} - \beta ( \mathbf{x}^{T}\mathbf{L}^{BE}\mathbf{x} ) - \lambda (\mathbf{x}^{T}\mathbf{x}  - 1) 
\end{equation}with $\beta \in \mathbb{R}$.  
Differentiating the Lagrangian and comparing the results to zero, we obtain the following generalized eigenvalue problem: 

\begin{equation}
\label{generalized eigenvalue problem}
( \mathbf{W}^{BE}   - \beta \mathbf{L}^{BE}) \mathbf{x} =  \lambda \mathbf{x}  
\end{equation}
where the eigenvector $\mathbf{x}$ solving (\ref{generalized eigenvalue problem}) is also a minimizer for (\ref{Feasiblility_Domain}). 

Denoting  
\begin{equation}
\label{Generlaized_Lap}
\mathbf{\tilde{L}}(\beta) =  \mathbf{W}^{BE}   -  \beta\mathbf{{L}}^{BE} 
\end{equation}we can write the generalized eigenvalue problem (\ref{generalized eigenvalue problem}) using the matrix pencil $\mathbf{\tilde{L}}(\beta)$ 
\begin{equation}
\mathbf{\tilde{L}}(\beta)\mathbf{x} =  \lambda \mathbf{x}  
\end{equation}
The scalar $\beta \in \mathbb{R}$ can be considered as a parameter that controls the trade-off between the total vertex and spectral spreads. An example for the trade-off in the barbell graph as captured by $\tilde{\mathbf{L}}^{BE}(\beta$)  is illustrated in Fig. \label{Eigenvectors_Barbell} (a) and (b) using $\beta$ = -200, and $\beta$ = -0.2, respectively. Functions colored with green correspond to the eigenvectors of $\tilde{\mathbf{L}}^{BE}(\beta$),  functions colored with blue correspond to the eigenvectors of ${\mathbf{L}}^{BE}$ .
As can be seen, when  $|\beta|$ is large,  the eigenvectors of $\tilde{\mathbf{L}}^{BE}(\beta$)  reveal structure which is similar to the eigenvectors of ${\mathbf{L}}^{BE}$, while small values of $|\beta|$ produce structure which is similar to those corresponding to ${\mathbf{W}}^{BE}$.

\subsection{Graph Representation and Interpretation}
For analysis and practical considerations it is often useful to encode the network captured in $\mathbf{\tilde{L}}$ using a semi positive definite operator. 
For practical consideration, we transform $\mathbf{\tilde{L}}$ into a semi positive definite matrix $\mathbf{L}_{\Delta}$ using a simple perturbation matrix $\Delta$, where $\Delta = \mu \mathbf{I} $, $\mu = - \tilde{\lambda}_{0} $, where $\tilde{\lambda}_{0} $ is the smallest eigenvalue of $\mathbf{\tilde{L}}$, and $\mathbf{I}
$ is the identity matrix. 
Setting 
 \begin{equation}
 \label{SPD_addition}
 \mathbf{L}_{\Delta} =   \mathbf{\tilde{L}} + \Delta 
 \end{equation}ensures that $\mathbf{L}_{\Delta}$ is semi positive definite (SPD) matrix. Letting $\left \{ \tilde{\lambda}_{i} \right \}_{i=1}^{N}$ and $\left \{ \lambda^{\Delta}_{i} \right \}_{i=1}^{N}$ be the eigenvalues corresponding to $\mathbf{\tilde{L}} $ and  $\mathbf{{L}}_{\Delta}$, respectively, we can see that the choice made in (\ref{SPD_addition}) ensures that the original spacing in eigenvalues of $ \mathbf{\tilde{L}} $ is preserved in $\mathbf{{L}}_{\Delta}$ $
 \lambda^{\Delta}_{i+1} - \lambda^{\Delta}_{i} = \tilde{\lambda}_{i+1} -  \tilde{\lambda}_{i}$. 
 The embedding method proposed using $ \mathbf{L}_{\Delta} $ is coined Graph Spectral Spread Embedding (GSSE).
 Note that there is a geometric interpretation which is related to the way $\mathbf{\tilde{L}} (\beta)$ was obtained from the lower boundary curve enclosing the feasibility domain  $\mathbb{D}_{ ( s_{\mathbf{L} },  s_{\mathbf{W}}  )  } $;  Setting instead another perturbation matrix $\Delta = \mu \mathbf{I} $  with $\mu = - \lambda_{N}^{W} $ and defining
\begin{equation}
q(\beta)  ={ \mbox{min}  } ( {\lambda } (\mathbf{L}_{\Delta} (\beta) ) )
\end{equation}
where ${ \mbox{min}  } ( {\lambda } (\mathbf{L}_{\Delta} (\beta) ) ) $ corresponds to the minimum eigenvalue of $\mathbf{L}_{\Delta} $ .
we have that 
\begin{equation}
 \mathbf{g}_{\mathbf{W} }(\mathbf{x})- \beta \mathbf{g}_{\mathbf{L}} (\mathbf{x})  \geq q(\beta)
\end{equation}which defines a half plane in $\mathbb{D}_{ ( s_{\mathbf{L} },  s_{\mathbf{W}}  )  }$. 

\textbf{Properties:} The feasibility domain $\mathbb{D}_{ (s_{\mathbf{L}^{BE} }, s_{\mathbf{W}^{BE} })  } $ is a bounded set since  $\forall  (s_{\mathbf{L}^{BE}}, s_{\mathbf{W}^{BE} }) \in \mathbb{D}_{ (s_{\mathbf{L}^{BE}}, s_{\mathbf{W}^{BE} })   } $ 
\begin{equation}
0  \leq s_{\mathbf{L}^{BE} } \leq  \lambda_{N}^{BE}  \, \, \mbox{and}\\
- \lambda_{N}^{W^{BE}} \leq s_{\mathbf{W}^{BE}}  \leq  \lambda_{N}^{W^{BE}}
\end{equation}
where $\lambda_{N}^{W^{BE}}$corresponds to the largest eigenvalue of the affinity graph. 
We provide quantitative analysis that bound the largest eigenvalue of the betweeness graph centrality, $\lambda_{N}^{W^{BE}}$ by the largest value of the vertex betweeness centrality of the graph (see Lemma 2 in the appendix). 

\begin{figure}
\centering
\includegraphics[width=.45\columnwidth]{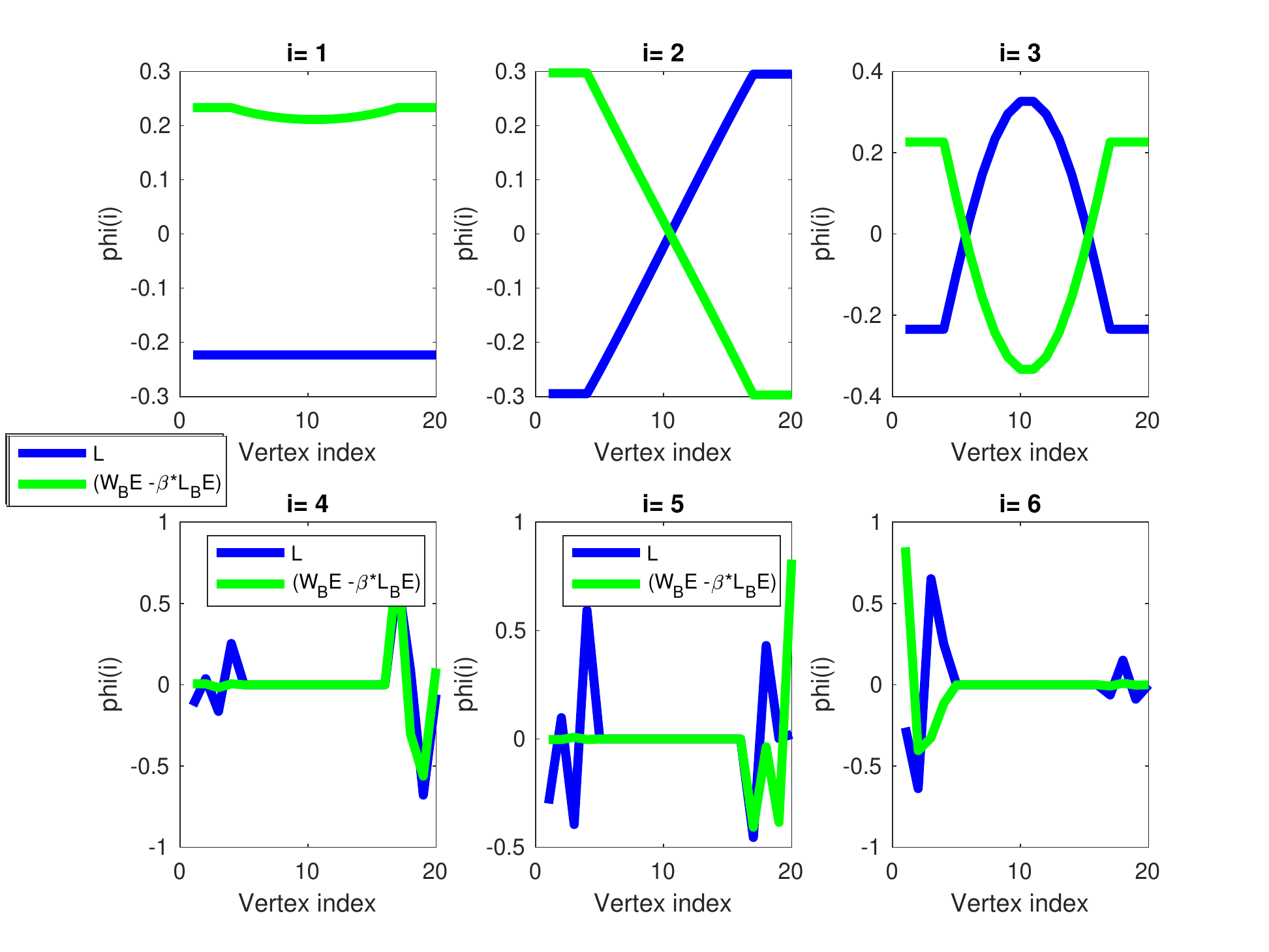}
\includegraphics[width=.45\columnwidth]{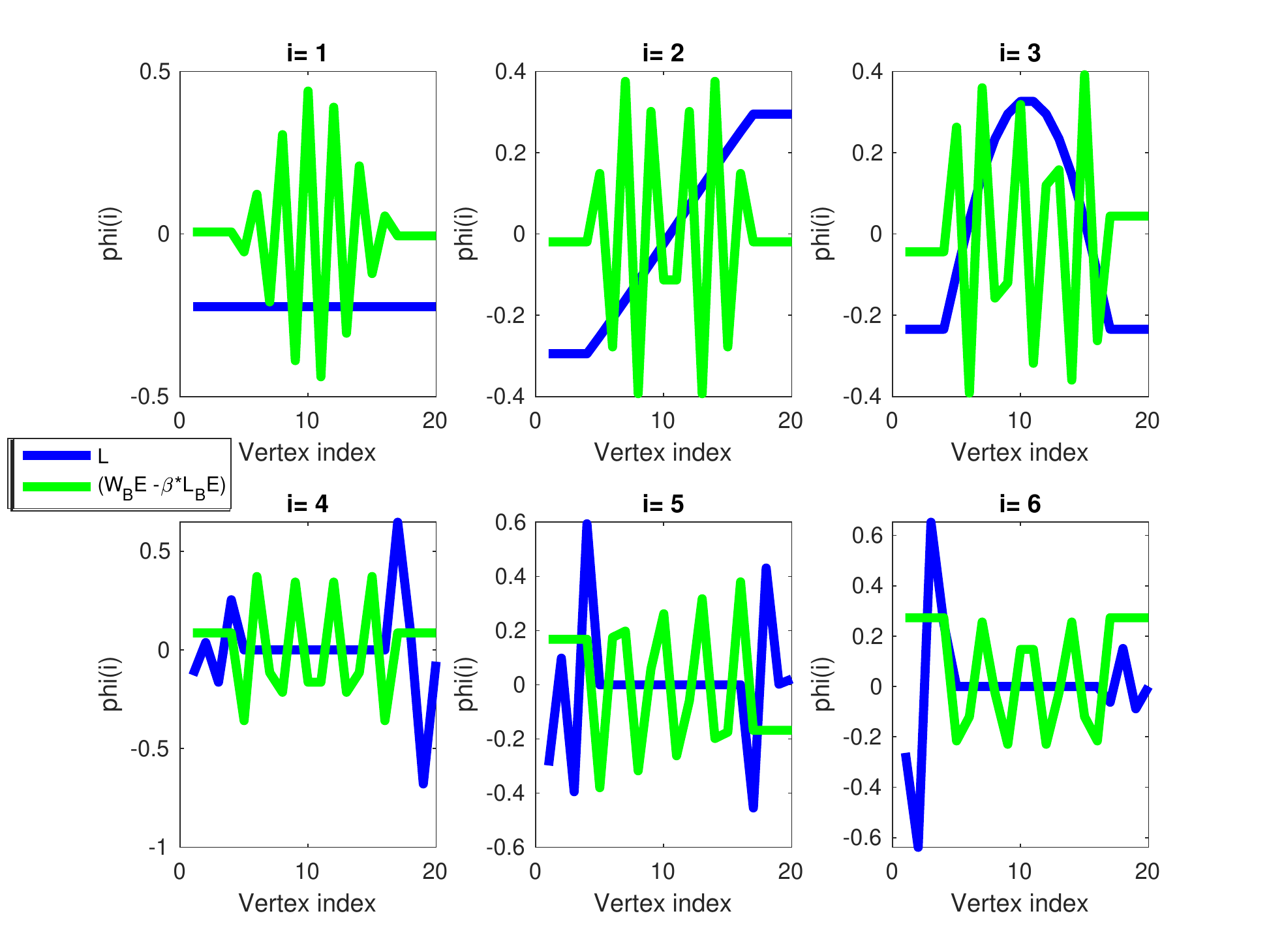}
\caption{The trade-off between local and structural node similarity in the barbell graph, as captured by $\tilde{\mathbf{L}}^{BE}(\beta$) in Figures (a) and (b), showing the eigenvectors associated with the smallest six eigenvalues corresponding to $\tilde{\mathbf{L}}^{BE}(\beta)$ (green color) using  $\beta$ = -200 in (a) and $\beta$ = -0.2 in (b).  Eigenvectors of $\mathbf{{L}}^{BE}$are shown in blue color. }

\label{Eigenvectors_Barbell}
\end{figure}

\subsection{Embedding using Sylvester equation}
\label{Embedding_Sylvester}
In this section we construct a node embedding using a linear mapping between the subspaces  $\mathbf{W}^{BE}$ and  $\mathbf{{L}}^{BE}$, which capture local and structural properties. We exploit the fact that the correspondence between the nodes represented by the two subspaces is known, and propose using the solution of a Sylvester equation as the node feature representation. This choice is motivated by Roth's removal rule, which states that one can find a complementary subspace by solving the Sylvester equation. $ \mathbf{W}^{BE}$and $\mathbf{{L}}^{BE}$ can be considered two different operators with known correspondence, the identity map. Thus, the resulting embedding given by the eigensystem of the Sylvester equation will be composed of two hybrid representations associated with the graph networks and local connectivity. The method is coined Graph Sylvester Embedding (GSE). 

The Sylvester operator $\mathbf{S}(\mathbf{X}) = \mathbf{A} \mathbf{X} + \mathbf{X} \mathbf{B}$  is used to express the eigenvalues and eigenvectors of $\mathbf{A}$ and $\mathbf{B}$ using a single operator $\mathbf{S}$, imposing commutativity, where we solve
\begin{equation}
\label{eqn:Sylvester}
\mathbf{S}(\mathbf{X})   =  \mathbf{C} 
\end{equation}
for $\mathbf{C} =  \rho \mathbf{I} $, a scalar regularizer, $\mathbf{I} $ is the identity matrix, where $\mathbf{L}^{BE}, \mathbf{W}^{BE},  \mathbf{C}  \in \mathbb{R}^{N\times N}$. 
The resulting eigensystem of $\mathbf{X}$ is described in the next Lemma below. 
\begin{Lemma}
\label{Eig_Sylvester}
Let $\mathbf{A}, \mathbf{B}$, and $ \mathbf{C} $ in the Sylvester equation  (\ref{eqn:Sylvester}), be symmetric. Assume that $ \mathbf{C}= \rho \mathbf{I}$ for some $\rho \in \mathbb{R}$. Let  $ \left \{u_{l}  \right \}_{l=1}^{N} $, $ \left \{  v_{l}  \right \}_{l=1}^{N} $, $ \left \{  \lambda_{l}   \right \} _{l=1}^{N}$, and  $ \left \{  \mu_{l}   \right \} _{l=1}^{N}$ the corresponding eigenvectors and eigenvalues of $\mathbf{A}$ and $  \mathbf{B}$, respectively. Moreover, assume that $\lambda_{l} \neq -\mu_{l}$ for all $l=1,..N$. Let $\mathbf{X}$ be the solution to the Sylvester system (\ref{eqn:Sylvester}). Then the associated eigenvalues and eigenvectors of $\mathbf{X}$ are $ \left \{ \frac{\rho}{\lambda_{l} +\mu_{l} }  \right \} _{l=1}^{N}$ and $ \left \{u_{l} +  v_{l}  \right \}_{l=1}^{N} $, respectively. 
\end{Lemma}

Applying Lemma \ref{Eig_Sylvester} with $ \mathbf{A}  = \mathbf{L}^{BE} $, and $ \mathbf{B}  = \mathbf{W}^{BE}$ we obtain that the set of eigenvectors corresponding to the solution $\mathbf{X}$ to the Sylvester equation \label{eq:Sylvester} are 
 $ \left \{\phi_{l}^{BE} + \phi_{l}^{W^{BE}}  \right \}_{l=1}^{N} $, where $\phi_{l}^{BE},\phi_{l}^{W^{BE}}$ correspond to the set of eigenvectors of $ \mathbf{L}^{BE} $, $ \mathbf{W}^{BE} $ , respectively. \\
The proposed embedding is constructed from the set of eigenvectors and associated eigenvalues corresponding to $ \mathbf{L}_{\Delta}^{BE}$ and $\mathbf{X}$, respectively. The performance of these embeddings is now evaluated empirically.

\section{Experimental Results}
\label{Experimental_Results}
We evaluate our method on both synthetic and real world networks in several applications including social, commerce and material science. We also report preliminary results on a recent dataset on protein interaction in Covid-19.\\
\textbf{Baseline:} To apply our approach in an unsupervised learning settings, we employ an off-the-shelf spectral descriptor, the Wave Kernel Descriptor (WKS) \cite{AubrySC11}. Given an input of  $r$ eigenvectors, their associated eigenvalues and a parameter $t$, the WKS descriptor constructs a feature vector $f(\mathbf{x}_{i}) \in \mathbb{R}^{t}$ for each node $\mathbf{x}_{i}$. To construct a graph embedding we compute the smallest $r$ eigenvalues of ${\mathbf{L}^{BE}_{\Delta}}$ and their associated eigenvectors and provide it as an input to the the WKS descriptor. Note that when using Sylvester embedding, we compute the largest $r$ eigenvalues and their associated eigenvectors corresponding to the solution $\mathbf{X} $.

The baseline is therefore the WKS descriptor applied to ${\mathbf{L}}$. As shown in the experimental results, the improvement in comparison to the baseline is up to 84\%.

\begin{figure}
\centering
\includegraphics[width=.2\columnwidth]{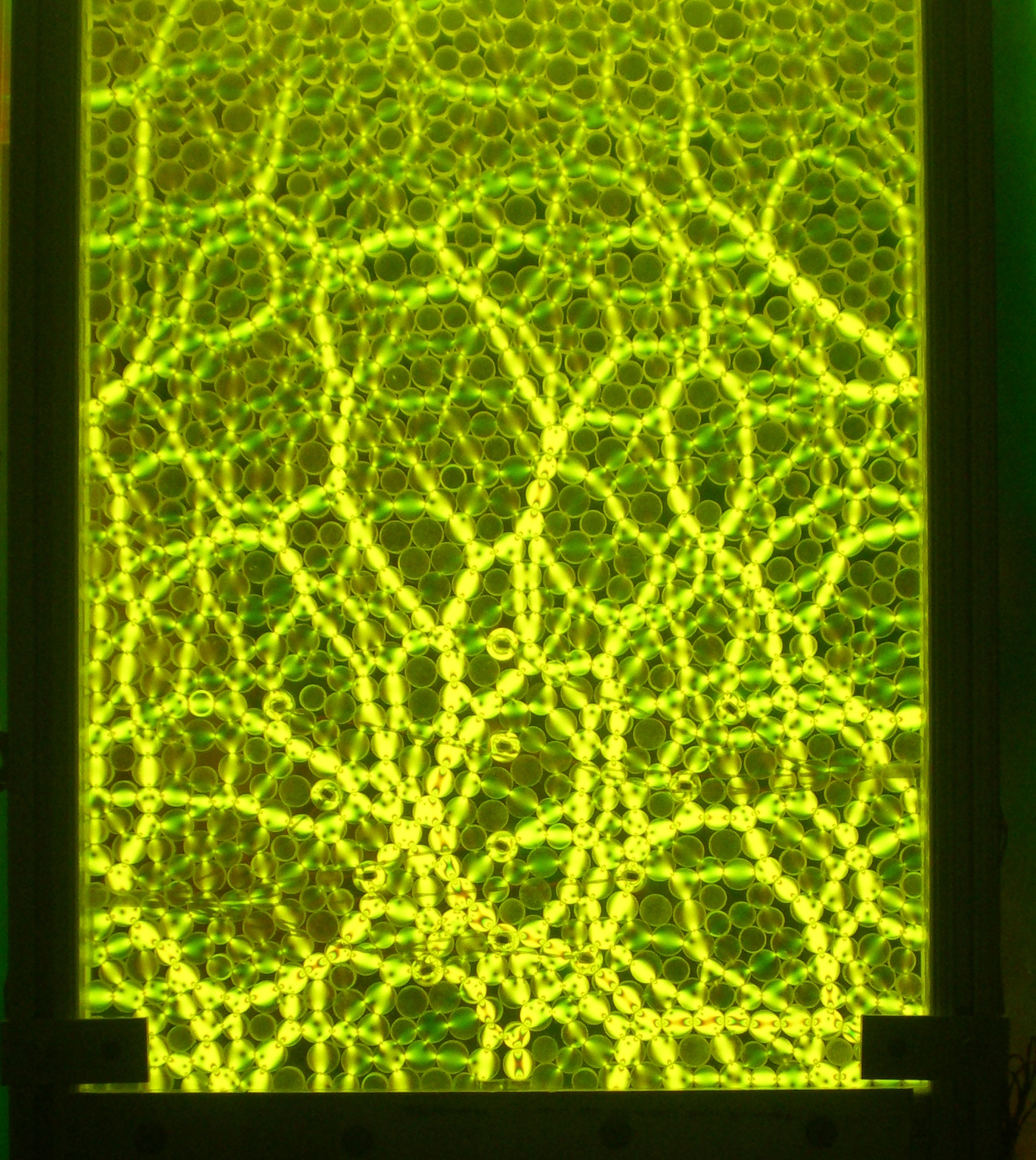}
\includegraphics[width=.35\columnwidth]{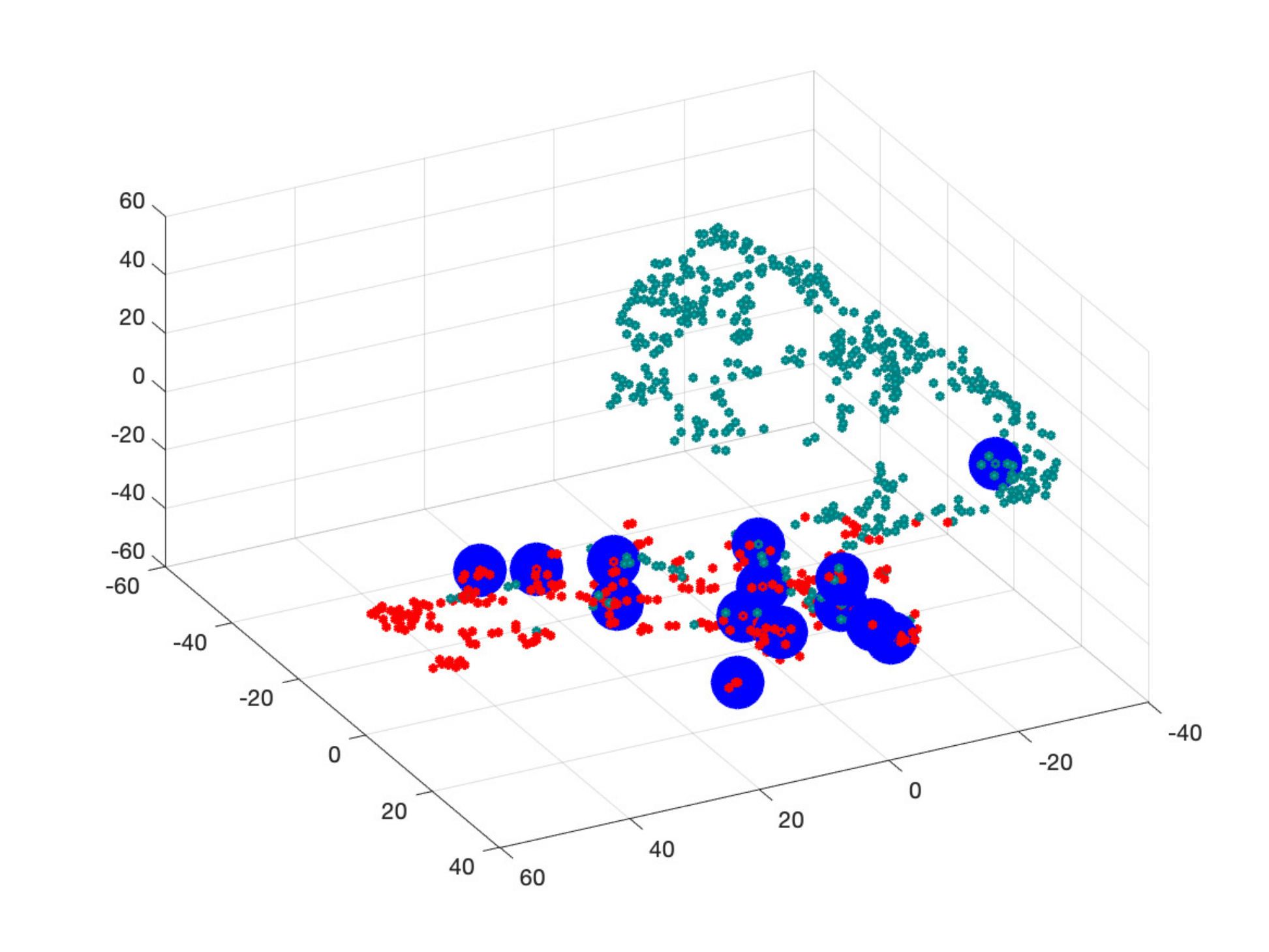}
\includegraphics[width=.35\columnwidth]{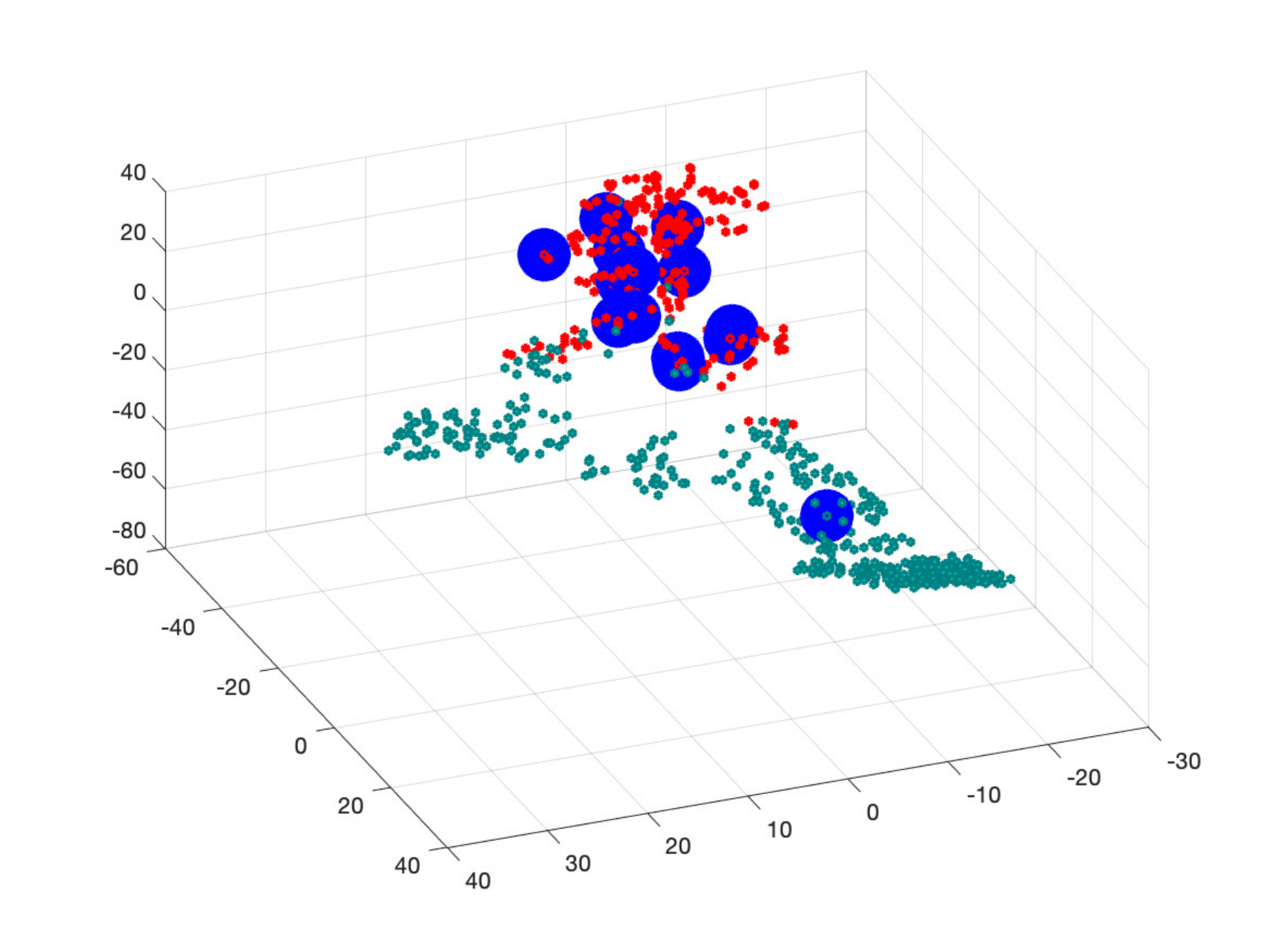}
\caption{Graph  embeddings used for forecasting failed edges using t-SNE embedding  (a) Contact network (yellow), which is extracted from the force chains recorded in a 2D assembly of frictional photoelastic disks overlaid on the reconstructed “pseudo-image" \cite{Berthier_2019}. (b) and (c) shows t-SNE visualization of edge embedding using the proposed Sylvester embedding and $\mathbf{L}^{BE}_{\Delta}$, respectively. Points with green color correspond to edges whose value is below the mean, and points with red color correspond to edges whose value is above the mean. The blue enlarged dots correspond to the failed edges in the system which were successfully detected by each method.}
  \label{Granualr_Embedding}
\end{figure}

\subsection{Node Classification}

\textbf{Datasets.} We compare our method using the Barbell graph (Figure \ref{fig:barbell_graph_illustrate}) for a qualitative comparison, and the mirrored Karate network for a quantitative comparison of node classification of structural similarity. The mirrored Karate network (based on a experiment suggested in \cite{Ribeiro_2017}) is composed of two copies corresponding to the Karate Club network \cite{Zac77}, where each node has a mirror node in the duplicated graph. We follow the exact same protocol which was suggested in \cite{Donnat_2018}, where the performance is evaluated by measuring the percentage of nodes whose nearest neighbor in the embedding space corresponds to its structural equivalent role.

\textbf{Barbell Graph} Our graph embeddings is able to identify nodes 5 and 16 as ones that stand out and have a large distance in the embeddings space from the rest of the nodes to which they are connected to (nodes 1 to 4 and 17 to 20, that are considered locally similar to 5 and 16, respectively). For example, the heat maps shown in Fig.\ref{fig:barbell_graph_illustrate}(b) and (c) represent embeddings using the same number of eigenvectors corresponding to $ {\mathbf{L}}$ and $ {\mathbf{L}^{BE}}$ respectively. In these cases, nodes 5 and 16 do not stands out and their distances to locally similar nodes is very small.  
\textbf{Karate network} We compare to a number of state-of-the-art methods in node feature embeddings including node2vec \cite{grover2016node2vec} struc2vec  \cite{Ribeiro_2017}, Deep Walk \cite{Deepwalk}, and GraphWave \cite{Donnat_2018}. The baseline which is the WKS descriptor  \cite{AubrySC11} combined with the Laplacian of the unweighted graph is also reported in Table \ref{tab:Mirrored_Karate_dataset}, where our approach shows a significant improvement over the baseline. The experimental results in Table \label{tab:Mirrored_Karate_dataset} (showing the best and average results for the varying number of edges) demonstrates that our method consistently performs well for a varying number of mirrored edges, and is improving the accuracy by more than $14 \%$ comparing to the best method.

\begin{table}[t]
\begin{center}
\tabcolsep=0.01cm
\begin{tabular}{|l|c|c|}
\hline
Method/Data&Best&Avg. \\
\hline
DeepWalk  \cite{Deepwalk}&  8.5 $\%$ &   8.5 $\%$   \\  
\hline
WKS  \cite{AubrySC11}  & 15.6 $\%$   &   \\
\hline
RolX \cite{RolX} & 84.5 $\%$ &  82.2 $\%$ \\
\hline
n2vec\cite{grover2016node2vec} &  8.5 $\%$  &  8.5 $\%$ \\
\hline
s2vec\cite{Ribeiro_2017} & 59.4 $\%$ & 52.5  \\
\hline 
GraphWave \cite{Donnat_2018}& 86 $\%$   &   83 $\%$ \\
\hline
\textbf{GSE} &\textbf{100  $\%$ } &  \textbf{88}  $\%$ \\
\hline
\textbf{GSSE} &\textbf{100  $\%$ }  & \textbf{88}  $\%$  \\
\hline
\end{tabular}
\end{center}
\caption{(Mirrored Karate Club: Percentage of nodes whose nearest neighbor corresponds to its structural equivalent nearest neighbor using our method compared to the state of the art methods in nodes feature representation on the mirrored karate club.} 
\label{tab:Mirrored_Karate_dataset}
\end{table}

\subsection{Node Classification on SARS-CoV-2 PPI Networks}
Effective representation of nodes with similar network topology plays an important role in the problem of protein protein interaction, where graph networks that describe protein-protein interaction (PPI) encode valuable information that can provide insights of biological function. 
To understand the development of COVID-19, the analysis of PPI networks is ongoing \cite{trendscovid}. It is important to develop robust computational methods that can reveal biological processes and host signaling pathways that may be used by SARS-CoV-2 to infect cells. 

One promising approach aims to develop drugs that target human proteins that the virus requires.  Recent research responding to COVID-19 aims to prioritize additional human protein by employing the known human protein interactors of SARS-CoV-2 proteins and a whole-genome protein interaction network in order to predict potential interactors with high accuracy \cite{law2020identifying}. 

Another important problem in studying PPI networks is the problem of network alignment or node correspondence which, by finding topological similarities between networks, can be useful to understanding biological function and evolution. 
We consider a problem of node classification by finding node correspondence using
network alignment on a recently published dataset of human proteins that physically interact with SARS-CoV-2 \cite{Gordon2020}. \\
\textbf{Dataset}: We test using the STRING network from the STRING database, a PPI network that consists of 18,886 nodes and 977,789 edges. The STRING network includes 330 human proteins that physically interact with SARS-CoV-2. It is likely to have both false positives and false negatives edges. For instance, the relative expression of the viral protein could increase the number of interacting partners detected, generating false positives  \cite{law2020identifying}. \\

We aim to find network alignment between two copies of the STRING network, given the STRING network and an additional STRING network which is created by randomly removing 10$ \%$ and 20 $\%$ percentage of noisy edges.
Each of the networks includes 330 nodes corresponding to the known human protein interactors of SARS-CoV-2. Given the node correspondence of the protein interactors of SARS-CoV-2, we connect each pair of nodes corresponding to the same human protein interactors of SARS-CoV-2 with an edge, which is resulted in a network composed of the two STRING networks. We then apply the proposed GSSE and GSE  to generate node embeddings. 
For comparison, we compute the node embeddings using graph embeddings methods including Laplacian Eigenmaps (LE) \cite{Belkin03},  Locally Linear Embedding (LLE)\cite{LLE}, Hope \cite{Hope}, and Regularized Laplacian (RL)\cite{Reg_Lap}, which are given the same input network. Additionally, we compare to network alignment methods, such as Isorank \cite{Isorank} and Final \cite{Final} and iNEAT \cite{iNeat}. For the network alignments methods, we provide as an input the corresponding affinity graphs and a matrix with the known correspondence between the nodes that correspond to human proteins that physically interact with SARS-CoV-2. 

Table \ref{tab:PPI_COVID} shows the percentage of nodes whose nearest neighbor corresponds to its true node correspondence using GSE and GSSE in the STRING network, which is compared to Graph Embeddings and network alignment methods. It is evident that graph embedding methods such as LLE and LE, which are rooted in manifold learning that is biased to local similarity and heavily relies on the graph smoothness are not effective for this task. Network alignments methods perform better, while our proposed GSE is improving the accuracy by more than $ \sim 16.4\%$ comparing to the best method.

\begin{table}[t]
\begin{center}
\tabcolsep=0.03cm
\begin{tabular}{|l|c|c|}
\hline
Method/noisy edges percentage & 10 $\%$ &  20 $\%$  \\
\hline
 LE \cite{Belkin03} &  5.38 $\%$ & 5.38  $\%$   \\  
\hline
LLE \cite{LLE}  & 1.5  $\%$   &  1.3 $\%$   \\
\hline
RL \cite{Reg_Lap} &  1$\%$ & -  \\
\hline 
HOPE  \cite {Hope}&  2 $\%$ &- \\
\hline
Isorank \cite{Isorank} & 41 $\%$  &  40 $\%$ \\
\hline 
Final \cite{Final} & 58.2 $\%$   & 56.6  $\%$ \\
\hline
iNEAT \cite{iNeat}  &  63.8 $\%$   & 56.1  $\%$ \\
\hline
\textbf{GSE} & \textbf{76.4$\%$}   & \textbf{60} $\%$ \\
\hline
\textbf{GSSE} &  48 $\%$  &  20.21 $\%$  \\
\hline
\end{tabular}
\end{center}
\caption{Node classification on the STRING network: percentage of nodes whose nearest neighbor corresponds to its true node correspondence using our method compared to Graph Embedding methods.} 
\label{tab:PPI_COVID}
\end{table}

\subsection{Forecasting Failed Edges}
Forecasting fracture locations in a progressively failing disordered structure is a crucial task in structural materials \cite{Berthier_2019}. Recently, networks were used to represent 2 dimensional (2D) disordered lattices and have been shown a promising ability to detect failures locations inside 2D disordered lattices. A promising aspect of this line of work is the possibility to assess failure locations of a structure without needing to study its detailed energetic states \cite{Berthier_2019}. In \cite{Berthier_2019}, it has been shown that the EBC can serve as a meaningful measurement to forecast the failure locations observed on a contact networks in 2D granular network. As failure locations were shown to occur predominately at locations that have large EBC values, \cite{Berthier_2019} proposed a thresholding scheme to compare the EBC of each edge to the mean EBC in order to forecast edges corresponding to failed beams. \\
\textbf{Dataset:} The set of disordered structures was derived from experimentally determined force networks in granular materials \cite{PhysRevMaterialsBerthier}. The network data is available in the Dryad repository \cite{PhysRevMaterialsBerthierDryad}. 
We tested 6 different initial networks, with mean degrees $z ={2.40, 2.55, 2.60, 3.35}$, 3.0, 3.6, following the same datasets corresponding to different initial granular configurations. 
For GSSE and GSE we used $\beta =-1000 $ and $\rho =1$, respectively, in all experiments. Table \ref{tab:Gran_network2} shows the success rates in forecasting failed edges for the 6 networks. In comparison to the forecasting approach proposed in [8], our framework performs consistently better. In particular, the proposed GSE which is based on Sylvester embedding provides the best results for all networks. 
We employ the tools developed in Sect. \ref{Properties_Spread}, with simple adaptation to create \textit{edge} embedding. First, by constructing node embedding, and then using the concatenated nodes features to construct edge embeddings. While our method is guided by the finding that failed edges occur predominantly in edges above the mean EBC, it doesn't rely on a single thersholding criteria to asses failed edges. Instead, we defer to make decisions about failing edges in the edge embedding space, which provides an effective tool to process, analyze and visualize the network. 
Fig. \ref{Granualr_Embedding} (b) and (c) are showing visualization using t-SNE of our proposed methods for edge embedding. The points colored in red correspond to the edges whose values are above the mean EBC, while the points colored in turquoise correspond to edges bellow the mean EBC. The blue enlarged dots correspond to the failed edges in the system which were successfully detected by each method.  In the examples illustrated, our proposed embeddeding successfully forecast all failed edges, which were mapped to the same cluster including edges whose EBC value was below the mean value. Table \ref{tab:Gran_network2} shows a quantitative results: the forecasting approach proposed \cite{Berthier_2019} will miss failed edges whose EBC is below the EBC mean, while embeddings using standard spectral embedding was found even less successful.

\section{Discussion}
\label{sec: Discussion}
We proposed a graph embedding to capture the  structure of a graph with both local and global characteristics, by incorporating basis functions that can be derived from the spectral representation of a graph coined the BCG. The BCG is obtained by modifying the original edges based on the edge betweeness centrality (EBC),  which measures the fractional number of shortest paths that are passing through an edge.  
Our work focuses on methods to learn the structure of graph networks without the need for supervision. This can be beneficial in performing data without the need for human annotation. Besides the cost of time and effort, this means that the data does not need to be seen by humans, which improves security and reduces the risk of accidental sharing of private data. 

We have proposed two different approaches: The first, GSSE, is rooted in the study of signal localization on irregular graphs, and it trades off  vertex and spectral domains to balance representations that capture local and structural similarity between the graph nodes.  The second, GSE is motivated by similar considerations is derived from solving a Sylvester equation. The advantage is that it allows more flexibility and control beyond scalar modulation between  the two bases.

Uncertainty principles inspire our work to incorporate the edge betweeness graph centrality (BCG) as a building block to create new graph topology that reflects node's structural role. 
In particular, the edge value in the edge betweeness graph centrality (EBC) is a useful tool to create an oscillatory pattern of higher frequencies which is also vertex localized. 
In contrast, unweighted graphs or graphs with very similar edges have limited ability to represent structural similarity as they associate signals decaying slowly with a larger graph neighborhood.
\subsection{Limitations and Future Work}
\textbf{Computational Complexity:} The implementation of our approach is computationally intensive, with most of the burden falling on the computation of the BCG, which is $O(mN)$ where $m$ is the total number of edges and $N$ is the total number of nodes, thus approximately $O(N^2) $ for sparse graphs. The execution time of our method with a Python code implementation using Intel Core i7 7700 Quad-Core 3.6GHz with 64B Memory on the Cora dataset with approximately 2700 nodes takes $\approx$  24.9 seconds.\\
\textbf{The local clustering coefficient:} Another limitation is concerned with applying our method to graphs in which the average local clustering coefficient is small, in particular when the graph contains a large number of vertices whose clustering coefficient is zero. 
If the degree of a node is 0 or 1, the corresponding clustering coefficient of that node is 0. In clustering tasks, our approach is shown to improves the baseline when the average local clustering coefficient is not too small, and is less effective when the average local clustering coefficient is very small.

Future work include expanding our approach to address semi-supervised applications, for example by incorporating generalized graph Laplacian or BCG in training using Graph convolutional networks. Another direction is to explore alternative approaches for finding a trade-off between local and structural similarity, tailored to a specific domain application using a functional applied to the adjacency graph in a similar spirit to the approach we propose with the BCG.

\vspace{-5pt}
\subsubsection*{Acknowledgments} 
This work was supported in part by ONR grant \#N00014-19-1-2229. The authors wish to thank Mason Porter for helpful discussions and suggestions in the earlier formulation of the manuscript. We also want to thank Yoram Louzoun, Roded Sharan, and Jie Chen for valuable feedback on the manuscript. We are also thankful for Simon Kasif, T. M. Murali, Mark Crovella, Catherine M. Della Santina, and Jeffrey Law for sharing insights from their work on Identifying Human Interactors of SARS-CoV-2 Proteins for COVID-19 using network label propagation, as well as sharing the STRING dataset.
\vspace{-5pt}

\begin{table*}
\centering
\begin{tabular}{|c|c|c|c|c|c|c|}
\hline
Method/ Network & z.2.40 & z.2.60  &  z.3.35 & z.2.55 & z.30 &  z.3.6 \\
\hline
FL  \cite{Berthier_2019}   & 85.7$\%$  &  70 $\%$  &   60$\%$ & 58.3  $\%$  & 51.5  $\%$ &  58.1$\%$ \\
\hline 
WKS \cite{AubrySC11} using $\mathbf{L}^{BE}$   & 50$\%$ &   80$\%$ &   56 $\%$   & 70.83  $\%$  &  60.0 $\%$ & 76.7 $\%$ \\
\hline
\textbf{GSE}  &\textbf{100 $\%$}   &\textbf{  85 $\%$}  &  \textbf{68$\%$} &\textbf{ 79} $\%$   & \textbf{78.7 $\%$} &\textbf{ 83.7$\%$} \\
\hline
\textbf{GSSE}  &\textbf{ 100 $\%$} & 80 $\%$   &  52 $\%$   & 70.83 $\%$ & 60.0 $\%$ & 76.7 $\%$ \\
\bottomrule
\end{tabular}
\caption{Success rate in forecasting failed edges in granular material networks}
\label{tab:Gran_network2}
\end{table*}


\section{Appendix}

The Sylvester operator $\mathbf{S}(\mathbf{X}) = \mathbf{A} \mathbf{X} + \mathbf{X} \mathbf{B}$  is used to express the eigenvalues and eigenvectors of $\mathbf{A}$ and $\mathbf{B}$ using a single operator $\mathbf{S}$, imposing commutativity, where we solve
\begin{equation}
\label{eqn:Sylvester}
\mathbf{S}(\mathbf{X})   =  \mathbf{C} 
\end{equation}
for $\mathbf{C} =  \rho \mathbf{I} $, a scalar regularizer, $\mathbf{I} $ is the identity matrix. We take  $ \mathbf{A}  = \mathbf{L}^{BE} $, and $ \mathbf{B}  = \mathbf{W}^{BE}$, where  $\mathbf{L}^{BE}, \mathbf{W}^{BE},  \mathbf{C}  \in \mathbb{R}^{N\times N}$. 
The resulting eigensystem of $\mathbf{X}$ is described in the next Lemma below. \\
\textbf{Lemma 1}
\label{Eig_Sylvester1}
\textit{Let $\mathbf{A}, \mathbf{B}$, and $ \mathbf{C} $ in the Sylvester equation  (\ref{eqn:Sylvester}), be symmetric. Assume that $ \mathbf{C}= \rho \mathbf{I}$ for some $\rho \in \mathbb{R}$. Let  $ \left \{u_{l}  \right \}_{l=1}^{N} $, $ \left \{  v_{l}  \right \}_{l=1}^{N} $, $ \left \{  \lambda_{l}   \right \} _{l=1}^{N}$, and  $ \left \{  \mu_{l}   \right \} _{l=1}^{N}$ the corresponding eigenvectors and eigenvalues of $\mathbf{A}$ and $  \mathbf{B}$, respectively. Moreover, assume that $\lambda_{l} \neq -\mu_{l}$ for all $l=1,..N$. Let $\mathbf{X}$ be the solution to the Sylvester system (\ref{eqn:Sylvester}). Then the associated eigenvalues and eigenvectors of $\mathbf{X}$ are $ \left \{ \frac{\rho}{\lambda_{l} +\mu_{l} }  \right \} _{l=1}^{N}$ and $ \left \{u_{l} +  v_{l}  \right \}_{l=1}^{N} $, respectively.} \\
\textbf{Proof of Lemma 1} Let $v_{l}$ be an arbitrary eigenvector of $\mathbf{B}$. Multiplying (\ref{eqn:Sylvester}) by $u_{l}$ we obtain 
\begin{equation}
( \mathbf{A} \mathbf{X} + \mathbf{X}  \mathbf{B} v_{l}) =  \mathbf{C} v_{l}
\end{equation}
since $v_{l} $ is an eigenvalue of $ \mathbf{A}  $ with associated eigenvalue $\mu_{l}$ we have
\begin{equation}
 \mathbf{A} \mathbf{X}  v_{l} + \mathbf{X} \mu_{l}  v_{l}  =  \mathbf{C} v_{l}
\end{equation}
Multiplying both sides of the equation above by $u_{l}^{T}$ we obtain 
\begin{equation}
u_{l}^{T}  \mathbf{A} \mathbf{X}  v_{l} + u_{l}^{T} \mathbf{X} \mu_{l}  v_{l}  = u_{l}^{T}   \mathbf{C} v_{l}
\end{equation}since $\mathbf{A}$ is assumed to be a symmetric matrix, we have that the left eigenvector  $u_{l}$ with corresponding eigenvalue $\lambda_{l}$ is also a right eigenvector with the same eigenvalue $\lambda_{l}$, hence 
\begin{equation}
u_{l}^{T} \lambda_{l} \mathbf{X}  v_{l} + u_{l}^{T} \mathbf{X} \mu_{l}  v_{l}  = u_{l}^{T}  \mathbf{C} v_{l}
\end{equation}Rearranging the equation above and dividing by $\lambda_{l} + \mu_{l}$ we obtain 
\begin{equation}
u_{l}^{T} \mathbf{X}  v_{l} =  \frac{ u_{l}^{T}   \mathbf{C} v_{l}  }  {   \lambda_{l} + \mu_{l}} 
\end{equation}
Next, note that $\mathbf{X}$ must be a symmetric matrix since $ \mathbf{A} $ and $ \mathbf{B}  $ are assumed to be symmetric matrices. Therefore, taking the transpose on of both size of the equation above we obtain: 
\begin{equation}
 v_{l}^{T} \mathbf{X} u_{l} =  \frac{ v_{l}^{T}  \mathbf{C}  u_{l}  }  {   \lambda_{l} + \mu_{l}} 
\end{equation}

thus 
\begin{equation}
(u_{l} +  v_{l})^{T}) \mathbf{X} (u_{l}+ v_{l}) =  \frac{ (u_{l} +  v_{l})^{T})   \mathbf{C}   (u_{l} +  v_{l})  }  {   \lambda_{l} + \mu_{l}} 
  \end{equation}
  and since $  \mathbf{C}= \rho \mathbf{I}$, then $ (u_{l} + v_{l})^{T}$ is an eigenvector of $\mathbf{X}$ with an associated eigenvalue $\frac{\rho } {\lambda_{l} + \mu_{l} } $ $\square$.  \\\\
The next Lemma, \textbf{Lemma 2} , states that the largest eigenvalue of the betweeness graph centrality, $\lambda_{N}^{W^{BE}}$, is bounded by the largest value of the vertex betweeness centrality of the graph. \\\\
\textbf{Lemma 2}
\label{Bound_betweeness_eigenvector}
\textit{Let $\sigma_{st}(i)$ denote the number of shortest distance paths passing between vertex $s$ and $t$ that are passing through a vertex $i \in \cal G$.  We have that the largest eigenvalue of the betweeness centrality graph satisfies: 
\begin{equation}
\lambda_{N}^{W^{BE}}  \leq \underset{i}{ \mbox{max}}  \sum_{s \neq t \neq i} \frac{ \sigma_{st} (i)} {\sigma_{st}}
\end{equation}}
\textbf{Proof of Lemma 2:} Let $\mathbf{x}, ||\mathbf{x}||= 1$ be the eigenvector associated with the largest eigenvalue $\lambda_{N}^{W^{BE}}   $of the geodesic edge betweeness centrality graph.
By definition of the betweeness centrality graph we have that 

\begin{align}
\lambda_{N}^{W^{BE}}  =\mathbf{x}^{T}  \mathbf{W}^{BE} \mathbf{x} = \sum_{i\sim j} w^{BE}_{i,j} x_{i}x_{j}   \leq  \underset{ i } {\mbox{max}} 
\sum_{i\sim j} \sum_{s \neq t }  \frac{ \sigma_{st} (i,j)} {\sigma_{st}} \\
\end{align}where the right hand is a result of the largest eigenvalue of a symmetric similarity matrix being bounded by the maximal vertex degree of the graph.
On the other hand, for any vertex $i \in G$ we have the equality 
\begin{equation}
\sum_{  \underset{  j }{i \sim j}  }   \sum_{s \neq t }   \frac{ \sigma_{st} (i,j)} {\sigma_{st}}  = \sum_{s \neq t \neq i} \frac{ \sigma_{st} (i)} {\sigma_{st}}
\end{equation}
and thus the proof is concluded. $\square$  \\
Note that the bound is not sharp for all graphs since in general there is no equality between the largest eigenvalue of a graph and the value of the node with maximal degree.  \\

\subsection{Additional experiments on Node Classification using  SARS-CoV-2 PPI Networks}

We report additional experimental results on the SARS- CoV-2 PPI networks for the task of node classification using network alignment. We follow a similar protocol as in the previous experiments that aim to find network alignment between two copies of the STRING network. An additional  STRING network is created by randomly removing 20$ \%$ of noisy edges from the STRING network. 
In this experiment, we randomly subsample the set of the 332 nodes corresponding to the known SARS-CoV-2 interactors and use only 163 nodes as the given known node correspondence. Since Graph Embeddings are found less effective for this task, we only compare to network alignments methods, including Isorank (Singh, Xu, and Berger 2008), Final (Zhang and Tong 2016), and iNEAT (Zhang et al. 2020). Table 1 shows the percentage of nodes whose nearest neighbor corresponds to its true node correspondence using GSE in the STRING network. The comparison results demonstrate that our method remains robust even when using a much smaller number of correspondence nodes.

\begin{table}[t]
\begin{center}
\tabcolsep=0.03cm
\begin{tabular}{|l|c|c|}
\hline
Method/noisy edges percentage & 20 $\%$  \\
\hline
Isorank (Singh, Xu, and Berger 2008) & 23 $\%$ \\
\hline 
Final  (Zhang and Tong 2016)   & 36  $\%$ \\
\hline
iNEAT (Zhang et al. 2020) & 37  $\%$ \\
\hline
\textbf{GSE} & \textbf{58$\%$}   \\
\hline
\end{tabular}
\end{center}
\caption{Node classification on the STRING network using 163 known SARS-CoV-2 interactors: percentage of nodes whose nearest neighbor corresponds to its true node correspondence using our method compared network alignments methods.} 
\label{tab:PPI_COVID}
\end{table}
\subsection{Clustering}
\textbf{Datasets.} The European air-traffic network \cite{Ribeiro_2017} is a data collected from the statistical office of the European union. The network has 399 nodes and 5,995 edges. Airport activity is measured by the total number of landings plus takeoffs in the corresponding period \cite{Ribeiro_2017}. The Cora dataset \cite{Cora} is a citation network where nodes are documents and edges are citation links. The network has 2,708 nodes and 5,429 edges. \textbf{Evaluation:} We evaluate our approach in clustering accuracy comparing to the ground truth using purity measure. We compute node feature representation for each node and then construct a new graph which is based on the $k=15$ nearest neighbor graph based on the new features. We compare our method to a number of clustering methods, including Spectral Clustering \cite{spectral-clustering} and Affinity Propagation. \\
\textbf{Summary of the experimental results:} on the 
\textbf{European air-traffic network} The experimental results, compare to spectral clustering and affinity propagation and measured in clustering accuracy by purity evaluation, results in an improvement of 24.1$\%$ and 28.2 $\%$ using GSSE, and 20$\%$ and 20.3 $\%$ using GSE, respectively, while using BCG directly we obtain an improvement of 6 $\%$ and 10$\%$.

\textbf{Citation network} For the Cora dataset, the performance comparing to spectral clustering and affinity propagation using the GSSE and GSE shows an improvements of less than $1 \%$, while using the BCG shows an improvement of 29.35$\%$ comparing to spectral clustering and affinity propagation.

\end{document}